\documentclass[10pt,twocolumn,letterpaper]{article}

\usepackage[pagenumbers]{cvpr} 

\usepackage{graphicx}
\usepackage{amsmath}
\usepackage{amssymb}
\usepackage{mathtools}
\usepackage{booktabs}
\usepackage{xcolor}
\usepackage{colortbl}
\usepackage{makecell}

\DeclarePairedDelimiter\floor{\lfloor}{\rfloor}

\usepackage[pagebackref,breaklinks,colorlinks]{hyperref}

\usepackage[capitalize]{cleveref}
\crefname{section}{Sec.}{Secs.}
\Crefname{section}{Section}{Sections}
\Crefname{table}{Table}{Tables}
\crefname{table}{Tab.}{Tabs.}

\def\cvprPaperID{10771} 
\def\confName{CVPR}
\def\confYear{2022}

\begin{document}

\title{PolyViT: Co-training Vision Transformers on Images, Videos and Audio}

\author{Valerii Likhosherstov\thanks{Equal contribution.}$^{~~1,2}$ \quad Anurag Arnab$^{*1}$ \quad Krzysztof Choromanski$^{1}$ \quad 
Mario Lučić$^{1}$ \\ \quad Yi Tay$^{1}$ \quad Adrian Weller$^{2,3}$ \quad Mostafa Dehghani$^{*1}$ 
\\
$^1$Google Research $^2$University of Cambridge $^3$Alan Turing Institute \\
\tt\small{\{vlikhosherstov, aarnab, dehghani\}@google.com}
}

\maketitle

\begin{abstract}

Can we train a single transformer model capable of processing multiple modalities and datasets, whilst sharing almost all of its learnable parameters?
We present PolyViT, a model trained on image, audio and video which answers this question. %
By co-training different tasks on a single modality, we are able to improve the accuracy of each individual task and achieve state-of-the-art results on 5 standard video- and audio-classification datasets. %
Co-training PolyViT on multiple modalities and tasks leads to a model that is even more parameter-efficient, and learns representations that generalize across multiple domains.
Moreover, we show that co-training is simple and practical to implement, as we do not need to tune hyperparameters for each combination of datasets, but can simply adapt those from standard, single-task training.

\end{abstract}

\section{Introduction}

Transformers~\cite{transformer} are a flexible family of neural sequence-to-sequence models. %
While it was originally designed for natural language processing, it has recently been adapted to 
a range of perception tasks, such as classification of images~\cite{vit}, video~\cite{vivit} and audio~\cite{gong2021ast}.
Despite recent advances across different domains and tasks, current state-of-the-art methods train a separate model with different model parameters for each task at hand.

\begin{figure*}[tb]
\centering
\includegraphics[width=0.95\linewidth]{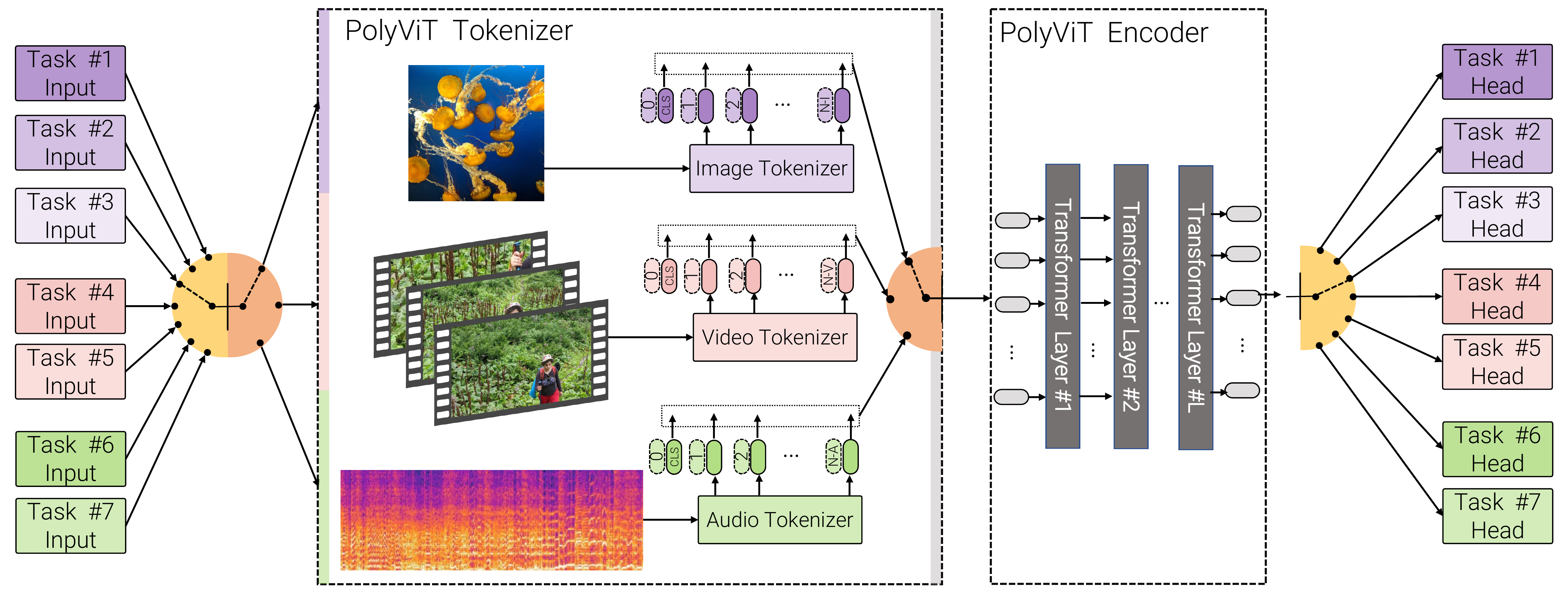}
\caption{Overview of PolyViT. 
Our model is capable of performing multiple tasks spanning different modalities, and processes a single task at a time.
The architecture consists of a transformer encoder shared among all tasks, modality-specific input tokenizers and task-specific %
output heads.
}
\vspace{-1\baselineskip}
\label{fig:teaser}
\end{figure*}

In this work, we present a simple yet effective method of training a single, unified model (Fig.~\ref{fig:teaser}) that achieves competitive, or state-of-the-art results for image-, video-, and audio-classification.
We go beyond using a common architecture for different modalities~\cite{jaegle2021perceiver}, as we also share model parameters across tasks and modalities, thus enabling potential synergies.
Our approach is motivated both technically, by the fact that transformers are generic architectures that can operate on any modality that can be tokenized, and intuitively, 
since human perception is inherently multimodal and performed by a single brain.

Our main technique is \emph{co-training}: training a single model on multiple classification tasks (across potentially multiple modalities) simultaneously.
We consider various settings, and simultaneously solve as many as 9 different image-, video- and audio-classification tasks.
As shown in Fig.~\ref{fig:teaser}, our model is capable of performing multiple tasks, but performs a single task at a time for a given input.
Although similar techniques have been explored in computer vision~\cite{maninis_cvpr_2019} and natural language~\cite{raffel2019exploring}, we are not aware of previous work that have considered multiple modalities and achieved state-of-the-art results with this approach.

We show that our co-training setup has multiple benefits:
In particular, it is parameter-efficient as we share the transformer parameters for each of the $n$ tasks of interest, approximately reducing the number of parameters by a factor of $n$. 
This has practical advantages when deploying models on edge devices (such as smartphones and embedded devices) with limited memory which may not otherwise be able to fit the weights of $n$ different models~\cite{iandola2017squeezenet}.
Furthermore, maintaining a single model for multiple tasks simplifies model deployment and online updates~\cite{iandola2017squeezenet}.
It is also noteworthy that co-training on tasks of the same modality leads to accuracy improvements on each individual task whilst also linearly decreasing total parameters. %
In particular, we achieve state-of-the-art results on video and audio classification across 5 different datasets.
This is facilitated by our observation that co-training has a regularizing effect, that improves performance on smaller datasets that large transformer models would otherwise overfit on.
In addition, when we extend co-training to multiple tasks and modalities, we observe that our accuracy is still competitive with the state-of-the-art whilst being even more parameter-efficient -- our model trained on 9 datasets uses 8.3 times fewer parameters whilst having at most a 1.2\% accuracy drop compared to state-of-the-art single-task baselines.
Finally, linear probing experiments show that this multi-task, multi-modal model is able to learn representations that generalize across multiple tasks and domains.
Once again, this has practical advantages when deploying models, as it shows that we can add new capabilities to the model by simply training an additional linear classifier. %

In addition to all the benefits outlined above, our co-training setup is simple and practical to implement.
It does not require hyperparameter tuning for each combination of co-training datasets, as we can readily adapt the settings of standard, single-task training.
In addition, co-training does not increase the overall training cost either, as the total number of training steps does not exceed that of the sum of each single-task baseline.
PolyViT is developed in Scenic~\cite{dehghani2021scenic}. For reproducibility, we will release code and checkpoints.

\section{Related work}

Our model is related to multi-task learning and transformer models, which we discuss below.

Multi-task learning aims to develop models that can address multiple tasks whilst sharing parameters and computation between them~\cite{caruana1997multitask}.
In computer vision, multiple papers have developed models which predict multiple outputs (for example semantic segmentation and surface normals), given a single input image~\cite{eigen2015predicting, kokkinos2017ubernet, zhang2014facial}.
Numerous works have also observed that although multi-task models are more versatile, their accuracies are lower than single-task models, and this accuracy deficit increases with the number of tasks, or by simultaneously performing unrelated tasks~\cite{kokkinos2017ubernet, zamir2018taskonomy, mccann2018natural}.
Moreover, jointly training a network to simultaneously perform multiple tasks has typically required careful calibration of the individual tasks, to ensure that none of the task-specific losses dominates another.
Methods to mitigate this include gradient-normalization~\cite{chen2018gradnorm} and -surgery~\cite{yu2020gradient} and adaptive loss weights~\cite{sener2018multi, kendall2018multi}. %

Our work differs in that although our network is capable of performing multiple tasks, it performs one task at a time for a given input. 
Note that this setting is also more suited to the case of handling multiple input modalities.
Such an approach was also taken by~\cite{maninis_cvpr_2019} who named it ``single-tasking of multiple tasks'' in the context of computer vision.
However, in natural language processing (NLP), this setting is still referred to as ``multi-task learning''~\cite{collobert2008unified}.
Furthermore, our co-training strategy is simple, and alternates between performing SGD for batches of separate tasks.
For high-capacity transformer models, we find that co-training on multiple datasets simultaneously helps to regularize the model on a dataset that it would otherwise overfit on, thus achieving accuracy improvements from co-training.
Previous works %
have improved performance on additional tasks only by introducing extra task-specific parameters~\cite{misra2016cross, houlsby2019parameter} which are typically conditioned on the input~\cite{rebuffi2017learning, maninis_cvpr_2019}.

We also note that similar co-training setups to our work have been explored in NLP.
A recent paradigm in NLP has been to reduce different tasks to a common, unified framework~\cite{raffel2019exploring, brown2020language, mccann2018natural}.
This common interface allows co-training a single model to perform multiple tasks, as it effectively involves concatenating multiple datasets together~\cite{raffel2019exploring, khashabi2020unifiedqa,tay2020hypergrid}.

Although the majority of previous multi-task learning works have considered only a single modality, Kaiser~\etal~\cite{kaiser2017one} presented an early effort on multi-modal models.
Their heterogeneous model consisted of convolutional layers to process images, and attention and mixture-of-experts layers to model text.
Their results, however, were not competitive with the state-of-the-art as ours.

Our model, motivated by \cite{vit}, %
can readily handle diverse modalities, as transformers operate on any sequence of tokens. %
Relevant to us, Perceiver~\cite{jaegle2021perceiver} is a transformer architecture that can process different modalities.
Instead of tokenizing images or audio spectrograms with non-overlapping patches like~\cite{vit} and~\cite{gong2021ast} respectively,~\cite{jaegle2021perceiver} operate directly on the raw input by projecting it into a smaller, latent set of tokens using cross-attention.
Although this architecture is capable of processing different modalities, the authors train separate networks with separate parameters for each task.
Therefore, they do not consider co-training scenarios like our work.
MBT~\cite{mbt}, on the other hand, proposes a transformer model to fuse different modalities (for example audio and rgb frames of videos) to solve a single task.
Once again, %
separate model parameters are used for each task.

UniT~\cite{hu2021unit} co-train a transformer-based model, but specifically for vision-and-language tasks.
The authors use an encoder-decoder architecture~\cite{transformer}, where only the decoder is shared among different tasks, and the encoder is specialized for each modality.
In particular, the visual encoder is DeTR~\cite{carion2020end} and the text encoder is BERT~\cite{bert}, and each component is pretrained separately.
In contrast to our work, they do not consider scenarios where the entire transformer backbone is shared among different tasks, nor do they thoroughly analyze how to co-train multiple tasks and modalities like our work.
Furthermore, their approach does not outperform single-task baselines as our work does.
Other papers concentrating on multi-task learning of vision-and-language tasks include~\cite{lu2019vilbert, li2020unicoder, lu2020twelve}.
On a separate track, Bain~\etal~\cite{bain2021frozen} use a single transformer encoder to process both images and videos for video-text retrieval.
However, their model is still trained on a single dataset and task, and the authors process images with the transformer as part of a complex, curriculum-learning strategy.
This is in contrast with our work which simultaneously trains a model for multiple tasks across different modalities.

Finally, we note that transformers have been used to process multiple modalities~\cite{akbari2021vatt, lee2020parameter} for cross-modal self-supervised learning~\cite{alayrac2020self, miech2020end}.
Lee~\etal~\cite{lee2020parameter} train a transformer on top of visual and audio features obtained from convolutional networks.
And to make model training computationally feasible, they perform low-rank approximations of the parameter matrices~\cite{sainath2013low, yang2015deep} to reduce the total number of model parameters.
These approaches are thus complementary to our work which shares almost all parameters in the model among different tasks.

\section{Preliminaries} \label{sec:prelim}

We define a \textit{modality} as the type of input processed by the network.
In this work, we consider images, audio, and video (specifically, the sequence of image frames in a video) as three separate modalities.
We perform classification as it is a fundamental problem whose solutions are often extended to more complex ones~\cite{girshick2014rich, he2017mask}.
By \textit{task}, we refer to a pair of input modality and a set of classes from which one or multiple classes are to be selected for a given input.
Each task corresponds directly to a dataset, for example, ImageNet-1K~\cite{imagenet} for image classification or Kinetics 400~\cite{kinetics} for video classification.

\subsection{Vision Transformers and extensions}
\label{sec:vit}
The Vision Transformer (ViT,~\cite{vit}) is a transformer-based architecture for image classification that closely follows~\cite{transformer}.
In contrast to language which is intuitively tokenized into words, ViT extracts tokens from the input image, $\mathbf{x}^\textsc{img} \in \mathbb{R}^{H \times W \times 3}$, by splitting it into $N = \floor{H / h} \times \floor{W / w}$ non-overlapping patches, $\mathbf{x}_1, \dots, \mathbf{x}_N \in \mathbb{R}^{h \times w \times 3}$.
Each patch, $x_i$, is then projected into a token $\mathbf{z}_i \in \mathbb{R}^d$ by a linear operator $\mathbf{E}$, $\mathbf{z}_i = \mathbf{E} \mathbf{x}_i$ (\textit{input embedding operator}).
All tokens are then concatenated into a sequence, which is also prepended with a learnable \textit{class token} $\mathbf{z}_{cls} \in \mathbb{R}^d$.
Learnable \textit{positional embeddings} $\mathbf{p} \in \mathbb{R}^{(N + 1) \times d}$ are also added to this sequence as the transformer is otherwise permutation invariant.
We denote this tokenization process as
\begin{equation}
    \mathbf{z}^0 = \begin{bmatrix} \mathbf{z}_{cls} & \mathbf{E} \mathbf{x}_1 & \dots & \mathbf{E} \mathbf{x}_N \end{bmatrix} + \mathbf{p}. \label{eq:vit1}
\end{equation}
Note that the linear operator $\mathbf{E}$ can also be seen as a 2D convolution with kernel of size $h \times w$ and strides $(h, w)$. 
The sequence of tokens, $\mathbf{z}$, is then processed by a transformer encoder, consisting of $L$ layers.
Each layer, $\ell$, is applied sequentially, and performs the transformations,
\begin{align}
    \mathbf{y}^\ell = \mathrm{MSA} \left(\mathrm{LN} \left(\mathbf{z}^{\ell - 1}\right)\right) + \mathbf{z}^{\ell - 1} \\
    \mathbf{z}^\ell = \mathrm{MLP} \left(\mathrm{LN} \left(\mathbf{y}^\ell\right)\right) + \mathbf{y}^\ell,
\end{align}
where $\mathrm{MSA}$ denotes multi-head self-attention~\cite{transformer}, $\mathrm{MLP}$ is a neural network with a single hidden layer and a GeLU nonlinearity \cite{gelu}, and $\mathrm{LN}$ denotes layer normalization \cite{ln}.

For a $C$-class classification problem, the class logits produced by the model are obtained by applying an output \textit{linear head} on the encoded classification token, $\mathbf{z}_{cls}^{L}$, as
\begin{equation}
    \mathbf{W}_{out} \mathbf{z}_{cls}^L + \mathbf{b}_{out} \in \mathbb{R}^{C}, \label{eq:vit4}
\end{equation}
where $\mathbf{W}_{out} \in \mathbb{R}^{C \times d}$ and $\mathbf{b}_{out} \in \mathbb{R}^C$ are the linear head's learnable parameters.

\paragraph{Extensions of ViT to audio and video}
The \textit{Audio Spectrogram %
Transformer} (AST)~\cite{gong2021ast} follows the same architecture as ViT, with the only difference that its inputs are log-mel spectrograms.
Spectrograms are image-like, time-frequency representations of audio, and can be tokenized like images. Moreover, the best AST model was initialized from ViT models pretrained on large image datasets.

\textit{Video Vision Transformers} (ViViT)~\cite{vivit} are an extension of ViT to video.
The authors proposed four model variants, and we consider the unfactorized version (Model 1 in~\cite{vivit}).
This model differs from ViT only in the input tokenization process, which it extends from 2D image patches to 3D spatio-temporal ``tubelets''. %
Namely, a video input $\mathbf{x}^\textsc{vid} \in \mathbb{R}^{F \times H \times W \times 3}$ is split into $N = \floor{F / f} \times \floor{H / h} \times \floor{W / w}$ non-overlapping tubelets $\mathbf{x}_1, \dots, \mathbf{x}_N \in \mathbb{R}^{f \times h \times w \times 3}$.
Following ViT, a linear operator $\mathbf{E}^\textsc{vid}$, which can be interpreted as a 3D convolution, projects $\{ \mathbf{x}_i \}$ into a  sequence of tokens $\{ z_i = \mathbf{E}^\textsc{vid} \mathbf{x}_i \in \mathbb{R}^d \}$, and computations (\ref{eq:vit1}-\ref{eq:vit4}) are repeated.

\paragraph{Initialization}
Finally, note that ViT, ViViT and AST all achieve their highest performance when pretrained on a large-scale dataset such as ImageNet-21K~\cite{imagenet} or JFT~\cite{jft}.
More specifically, ViT was initially pretrained on ImageNet-21K or JFT, and then finetuned at higher resolution on target datasets such as ImageNet-1K. %
ViViT and AST also initialize their models from large-scale, image-pretrained models. %
In all of these cases, the positional embeddings, $\mathbf{p}$, which depend on the sequence length $N$ (and thus the input resolution), are interpolated from the pretrained model to the finetuned model.
Furthermore, the 3D embedding projection of ViViT, $\mathbf{E}^\textsc{vid}$, is initialized from the 2D projection of ViT, $\mathbf{E}^{\textsc{img}}$~\cite{vivit}.

The similarities between ViT, ViViT and AST allow us to construct a multi-modal model with a shared transformer encoder, and separate input tokenizers as described next. %

\section{Co-training ViT on images, audio and video}
\label{sec:polyvit}

\subsection{PolyViT architecture}

PolyViT is a single architecture that is capable of processing inputs from multiple modalities.
As shown in Fig.~\ref{fig:teaser}, we share a transformer encoder among different tasks and modalities, enabling up to a linear reduction in parameters with the number of tasks.
Note that PolyViT with $L$ layers acts like an $L$-layer ViT when processing images, an $L$-layer AST when processing audio, and an $L$-layer unfactorized ViViT when processing video.
And whilst it is capable of handling multiple modalities, it performs one task from one modality in a given forward pass.

As shown in Fig.~\ref{fig:teaser}, PolyViT employs modality-specific class tokens, $\mathbf{z}_{cls}^{\textsc{img}}, \mathbf{z}_{cls}^{\textsc{vid}}, \mathbf{z}_{cls}^{\textsc{aud}}$, input embedding operators, $\mathbf{E}^{\textsc{img}}, \mathbf{E}^{\textsc{vid}}, \mathbf{E}^{\textsc{aud}}$, and positional embeddings $\mathbf{p}^{\textsc{img}}, \mathbf{p}^{\textsc{vid}}, \mathbf{p}^{\textsc{aud}}$.
This allows the network to encode modality-specific information that can be leveraged by the subsequent, shared transformer backbone.
It also accounts for the fact that the number of tokens per modality may vary.

A separate output linear head (Eq.~\ref{eq:vit4}) is then used for each task with the following learnable weights:
\begin{equation}
    \mathbf{W}_{out} = \mathbf{W}_{out}^{mod,j} \in \mathbb{R}^{C_j \times d}, \ \mathbf{b}_{out} = \mathbf{b}_{out}^{mod,j} \in \mathbb{R}^{C^j}, \label{eq:theads}
\end{equation}
where $mod \in \{ \textsc{img}, \textsc{vid}, \textsc{aud} \}$ is a modality, $j \in \{ 1, \dots, T^{mod} \}$ is a task index in the set of tasks for that modality, $T^{mod}$ is the number of tasks for that modality and $C_j$ is the number of classes for that task.
Note that the output heads are the only task-specific parameters.
The input embedding operators, positional embeddings and class tokens are shared by all tasks within a modality.

To increase model capacity when co-training on a large number of tasks and modalities simultaneously, we can optionally include $L_{adapt} \geq 0$ modality-specific transformer layers (which we denote as \textit{modality-adaptor layers}).
These transformer layers are applied directly after tokenization.
In this case, there are $L_{shared} = L - L_{adapt}$ layers which are shared among all modalities and tasks.
We can think of this case as using a shallower transformer encoder, but a deeper subnetwork to extract tokens from each modality.

As almost all computation and parameters within our architecture are within the $L$ layers of the transformer encoder, if there are $n$ tasks, we reduce the total number of parameters by a factor of approximately $n$ when $L_{shared} = L$.
This is in comparison to standard, single-task training.
Note that the overall inference time does not change, as PolyViT still performs one task per forward pass.

\subsection{Co-training procedure} \label{sec:sch}
We optimize all PolyViT model parameters, $\theta$, simultaneously across all the tasks that we are co-training on with stochastic gradient descent (SGD).
As a result, there are a myriad of design choices on how to construct training batches, compute gradients to update model parameters, and which training hyperparameters to use.

In all cases, we construct our training minibatches using examples from a single task.
This design choice allows us to evaluate gradients and perform a parameter update using the same training hyperparameters (e.g., learning rate, batch size, and momentum) as a conventional single-task baseline.
As a result, we can perform co-training on multiple tasks without any additional hyperparameter tuning compared to the single-task baseline~\cite{vit, gong2021ast, vivit}, making co-training simple to perform in practice, and alleviating the need to perform large hyperparameter sweeps in order to achieve competitive accuracy.
Note that without this property, we would need to tune training hyperparameters on the product set of all co-training datasets, which would be computationally infeasible.
Constructing minibatches from a single task (where each example has the same number of tokens) has further computational advantages on GPU- or TPU-accelerators, as tokens do not need to be padded to a maximum sequence length.

\begin{figure}%
\centering
\includegraphics[width=\linewidth]{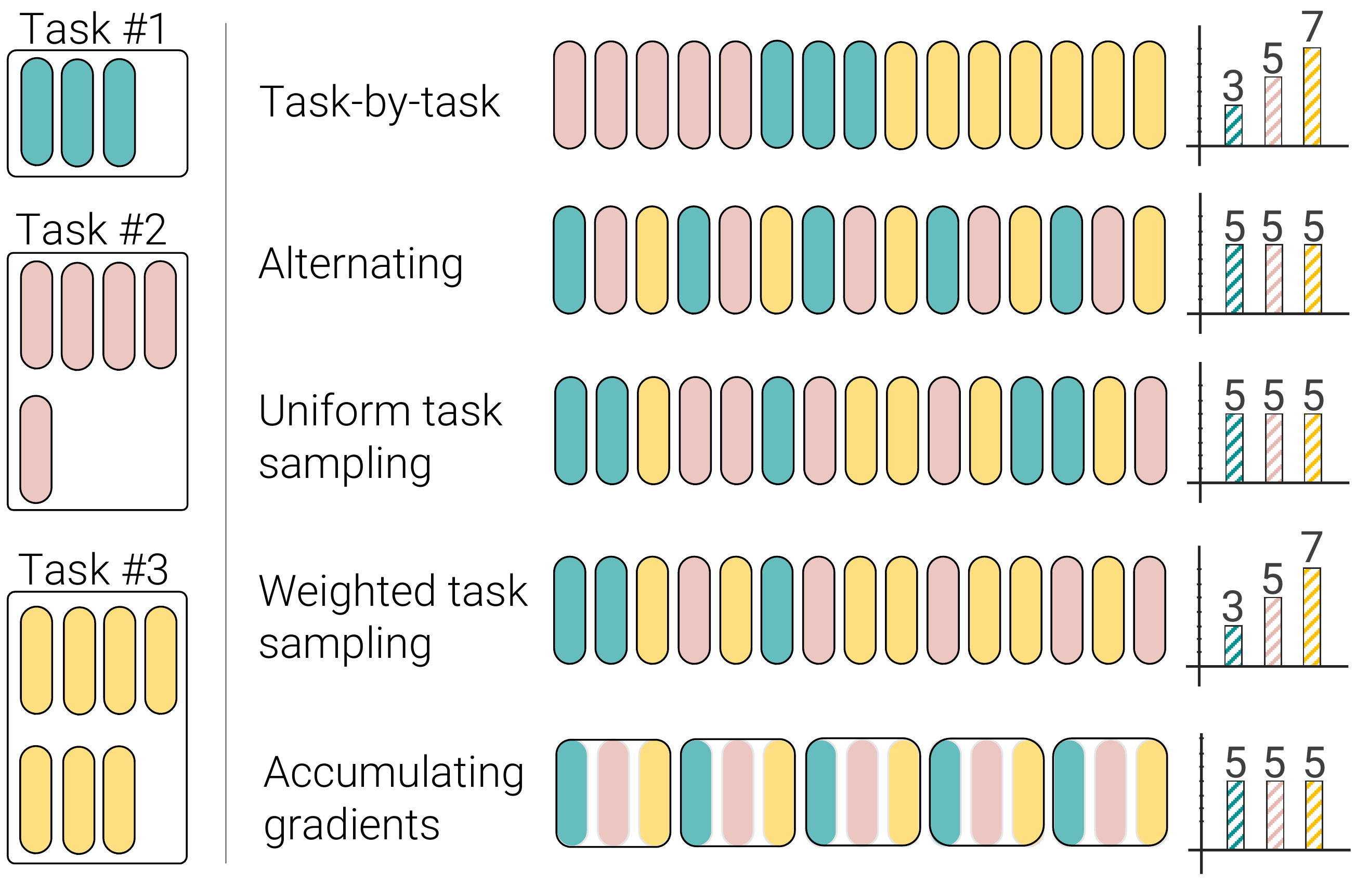}
\vspace{-1\baselineskip}
\caption{
Task sampling schedules considered in this paper.
Each element within a task corresponds to the number of training steps performed for that task by the baseline model.
}
\vspace{-1\baselineskip}
\label{fig:task_sampling_schedule}
\end{figure}

During co-training, for each SGD step, we sample a task (dataset), then sample a minibatch from that task, evaluate a gradient and then perform a parameter update.
An important consideration is the order in which we sample tasks and whether we accumulate gradients over different minibatches and tasks.
We describe several task sampling schedules below and in Fig.~\ref{fig:task_sampling_schedule}.
We first denote $U_j$ as the number of SGD steps for the single-task baseline that the original authors reported for their best model, where $j \in \{1, \ldots, T\}$ indexes the task and $T = T^{\textsc{img}} + T^{\textsc{aud}} + T^{\textsc{vid}}$ is the total number of tasks. %
Furthermore, we define $U$ as the total number of SGD steps during co-training.

\vspace{-0.5\baselineskip}
\paragraph{Task-by-task} %
In this schedule, the first $U_{j_1}$ SGD steps are performed with task $j_1$, the next $U_{j_2}$ steps using task $j_2$ and so on, where $[ j_1, \dots, j_T ]$ is a random task order.

\vspace{-0.5\baselineskip}
\paragraph{Alternating} This deterministic schedule alternates between tasks in a fixed, repeating order.
Concretely, we perform a single SGD step for each task in sequence before repeating the same order.
We set $U = \sum_{j = 1}^M U_j$ which implies $U / T$ training steps per task.

\vspace{-0.5\baselineskip}
\paragraph{Uniform task sampling}
This is a stochastic version of the schedule above, where the task for each SGD step is sampled from a uniform distribution, with probability $1 / T$.
We implement it such that the number of training steps for task $j$ is exactly $U / T$, by randomly permuting an array with $U$ elements, where $U / T$ elements correspond to each task.

\vspace{-0.5\baselineskip}
\paragraph{Weighted task sampling} 
In this stochastic schedule, we sample each task with a weight proportional to the number of training steps in the single-task baseline. %
Therefore, $U = \sum_{j = 1}^M U_j$, and the sampling weight for task $j$ is $U_j / U$.
We implement this schedule as above, to ensure that we perform exactly $U_j$ steps for task $j$.

\vspace{-0.5\baselineskip}
\paragraph{Accumulating gradients} For $T$ tasks, we perform a forward and backward pass on a minibatch for each task, summing the gradients over each task.
We then perform a single parameter update with the accumulated gradients, thus effectively using a larger batch size encompassing all the tasks being co-trained.
Here, we set $U = (\sum_{j = 1}^T U_j) / T$.

\subsection{Initialization of PolyViT}
As described in Sec.~\ref{sec:vit}, ViT, ViViT and AST models are initialized from models pretrained on ImageNet-21K or JFT before being finetuned for the task of interest.
In all of our experiments, we also finetune from a ViT model pretrained on ImageNet-21K, and follow the initialization methods for the positional embeddings, $\mathbf{p}$, and input embeddings, $\mathbf{E}$, for each modality as described in~\cite{vit} and~\cite{vivit}.

When we use modality-adaptor layers, that is $L_{adapt} > 0$, the first $L_{adapt}$ layers for each modality are initialized with the same first $L_{adapt}$ layers from the pretrained ViT model.
These parameters are however allowed to change from each other during training.
Similarly, shared PolyViT layers are initialized from the last $L_{shared}$ transformer encoder layers from the pretrained ViT model.
Note that the output linear heads are all initialized randomly.  %

\section{Experiments}
\label{sec:experiments}

\subsection{Experimental Setup} \label{sec:expsetup}

We train
PolyViT simultaneously on 9 diverse classification tasks spanning the image, video, and audio modalities.
Note that datasets and tasks have a one-to-one correspondence.
We chose this setup of 9 tasks, as the datasets include a large variation in domain and training set sizes.
Furthermore, the single-task baseline training hyperparameters vary substantially between the tasks.
Consequently, we believe this presents a challenging co-training setup.

When co-training for image classification, we use ImageNet-1K, CIFAR-10 and -100, Oxford-IIIT Pets, and RESISC45.
For video, we use Kinetics 400 and Moments in Time, and for audio, AudioSet and VGGSound.
Exhaustive details of these datasets are in Appendix~\ref{sec:addexp}.
As in~\cite{mbt}, we evaluate on the whole AudioSet validation set, but use a smaller balanced subset referred to Mini-AudioSet (MiniAS) for initial experiments.
We then use a larger, balanced subset of 500 000 examples (referred to AS-500k) for our state-of-the-art comparisons following~\cite{mbt}.
We follow standard evaluation protocols for each task, reporting classification accuracy (\%) for all tasks except  AudioSet, where we report mean average precision (mAP) as it is multilabel. %

\begin{table*}[t]
\centering
\caption{The effect of the task sampling schedule on co-training performance on multiple modalities and tasks.
The highest accuracy is shown in bold, and the second-highest is underlined.
Note how the ``Weighted'' task sampling method consistently achieves the highest accuracy for 8 out of 9 tasks, and second-highest on the remainder.
Results are on the validation set. 
}
\label{tab:sch}
\scalebox{0.9}{
\begin{tabular}{l ccccc cc cc}
\toprule
         & \multicolumn{5}{c}{Image}      & \multicolumn{2}{c}{Video} & \multicolumn{2}{c}{Audio} \\
         \cmidrule(lr){2-6} \cmidrule(lr){7-8} \cmidrule(l){9-10}
Schedule & Im1K &  C100 & C10 & Pets & R45 & K400         & MiT        & MiniAS        & VGG       \\
\cmidrule(r){1-1} \cmidrule(lr){2-6} \cmidrule(lr){7-8} \cmidrule(l){9-10}

Task-by-task & \cellcolor[rgb]{1.0, 0.6, 0.6} 0.3 & \cellcolor[rgb]{1.0, 0.6, 0.6} 0.8 & \cellcolor[rgb]{1.0, 0.6, 0.6} 11.7 & \cellcolor[rgb]{1.0, 0.6, 0.6} 1.9 & \cellcolor[rgb]{1.0, 0.6, 0.6} 2.0 & \cellcolor[rgb]{1.0, 0.6, 0.6} 0.3 & \cellcolor[rgb]{1.0, 0.6, 0.6} 0.3 & \cellcolor[rgb]{1.0, 0.6, 0.6} 1.6 & \cellcolor[rgb]{1.0, 0.96, 0.33} 37.2\\
Accumulated & \cellcolor[rgb]{0.5, 1.0, 0.5} \textbf{88.1} & \cellcolor[rgb]{0.5, 1.0, 0.5} \underline{90.0} & \cellcolor[rgb]{0.51, 1.0, 0.5} 98.8 & \cellcolor[rgb]{0.53, 1.0, 0.49} 94.0 & \cellcolor[rgb]{0.51, 1.0, 0.5} 96.1 & \cellcolor[rgb]{0.69, 1.0, 0.42} 58.0 & \cellcolor[rgb]{0.81, 1.0, 0.38} 22.5 & \cellcolor[rgb]{0.83, 1.0, 0.37} 22.9 & \cellcolor[rgb]{1.0, 0.6, 0.6} 27.3\\
Alternating & \cellcolor[rgb]{0.52, 1.0, 0.49} 86.0 & \cellcolor[rgb]{0.51, 1.0, 0.5} 89.4 & \cellcolor[rgb]{0.5, 1.0, 0.5} \underline{99.2} & \cellcolor[rgb]{0.53, 1.0, 0.49} 94.0 & \cellcolor[rgb]{0.51, 1.0, 0.49} 95.8 & \cellcolor[rgb]{0.53, 1.0, 0.49} \underline{69.7} & \cellcolor[rgb]{0.58, 1.0, 0.47} \underline{30.0} & \cellcolor[rgb]{0.57, 1.0, 0.47} \underline{31.4} & \cellcolor[rgb]{0.71, 1.0, 0.42} \underline{44.6}\\
Uniform & \cellcolor[rgb]{0.53, 1.0, 0.49} 85.8 & \cellcolor[rgb]{0.51, 1.0, 0.5} 89.3 & \cellcolor[rgb]{0.51, 1.0, 0.5} 98.6 & \cellcolor[rgb]{0.52, 1.0, 0.49} \underline{94.6} & \cellcolor[rgb]{0.51, 1.0, 0.5} 96.1 & \cellcolor[rgb]{0.54, 1.0, 0.48} 68.8 & \cellcolor[rgb]{0.6, 1.0, 0.46} 29.3 & \cellcolor[rgb]{0.59, 1.0, 0.46} 30.6 & \cellcolor[rgb]{0.73, 1.0, 0.41} 44.1\\
Weighted & \cellcolor[rgb]{0.51, 1.0, 0.49} \underline{86.9} & \cellcolor[rgb]{0.5, 1.0, 0.5} \textbf{90.4} & \cellcolor[rgb]{0.5, 1.0, 0.5} \textbf{99.3} & \cellcolor[rgb]{0.5, 1.0, 0.5} \textbf{96.5} & \cellcolor[rgb]{0.5, 1.0, 0.5} \textbf{97.0} & \cellcolor[rgb]{0.5, 1.0, 0.5} \textbf{71.6} & \cellcolor[rgb]{0.5, 1.0, 0.5} \textbf{32.5} & \cellcolor[rgb]{0.5, 1.0, 0.5} \textbf{33.5} & \cellcolor[rgb]{0.5, 1.0, 0.5} \textbf{49.2}\\
\bottomrule
\end{tabular}
}
\end{table*}
\begin{table*}[t]
\caption{Co-training with PolyViT-Base. As indicated by the ``\#Models'' column, some rows correspond to multiple trained models. In this case, we report the total number of parameters across all the models.
PolyViT co-trained on a single-modality outperforms single-task baselines in most cases, whereas PolyViT co-trained on multiple modalities achieves competitive performance with a large reduction in parameters.
Results are on the test set.
Further dataset details in Appendix~\ref{sec:addexp}.
}
\label{tab:cotr}
\begin{center}
\vspace{-1\baselineskip}
\scalebox{0.76}{
    \begin{tabular}{l cc ccccc cc cccc}
    \toprule
             &  & & \multicolumn{5}{c}{Image}      & \multicolumn{2}{c}{Video} & \multicolumn{2}{c}{Audio} \\
             \cmidrule(lr){4-8} \cmidrule(lr){9-10} \cmidrule(l){11-12}
    Model & \#Models & \#Params &  Im1K & C100 & C10 & Pets & R45 & K400         & MiT        & MiniAS        & VGG \\
    \cmidrule(r){1-1} \cmidrule(lr){2-3} \cmidrule(lr){4-8} \cmidrule(lr){9-10} \cmidrule(l){11-12}

ViT-Im21K Linear probe & 1 & \textbf{93M} & \cellcolor[rgb]{1.0, 0.6, 0.6} 80.7 & \cellcolor[rgb]{1.0, 0.6, 0.6} 76.2 & \cellcolor[rgb]{1.0, 0.6, 0.6} 91.7 & \cellcolor[rgb]{1.0, 0.6, 0.6} 91.8 & \cellcolor[rgb]{1.0, 0.6, 0.6} 81.7 & \cellcolor[rgb]{1.0, 0.6, 0.6} 64.0 & \cellcolor[rgb]{1.0, 0.6, 0.6} 25.5 & \cellcolor[rgb]{1.0, 0.6, 0.6} 11.3 & \cellcolor[rgb]{1.0, 0.6, 0.6} 15.7\\
Single-task baseline & 9 & 773M & \cellcolor[rgb]{0.83, 1.0, 0.37} 83.1 & \cellcolor[rgb]{0.58, 1.0, 0.47} 92.0 & \cellcolor[rgb]{0.51, 1.0, 0.49} 99.0 & \cellcolor[rgb]{0.68, 1.0, 0.43} 94.5 & \cellcolor[rgb]{0.5, 1.0, 0.5} \textbf{96.7} & \cellcolor[rgb]{0.59, 1.0, 0.46} 78.7 & \cellcolor[rgb]{0.75, 1.0, 0.4} 33.8 & \cellcolor[rgb]{0.79, 1.0, 0.38} 29.3 & \cellcolor[rgb]{0.5, 1.0, 0.5} \textbf{51.7}\\
\cmidrule(lr){1-12} PolyViT, 1 modality & 3 & 263M & \cellcolor[rgb]{0.5, 1.0, 0.5} \textbf{84.3} & \cellcolor[rgb]{0.5, 1.0, 0.5} \textbf{93.3} & \cellcolor[rgb]{0.5, 1.0, 0.5} \textbf{99.1} & \cellcolor[rgb]{0.5, 1.0, 0.5} \textbf{95.1} & \cellcolor[rgb]{0.52, 1.0, 0.49} 96.4 & \cellcolor[rgb]{0.5, 1.0, 0.5} \textbf{80.2} & \cellcolor[rgb]{0.5, 1.0, 0.5} \textbf{36.5} & \cellcolor[rgb]{0.5, 1.0, 0.5} \textbf{36.7} & \cellcolor[rgb]{0.5, 1.0, 0.5} 51.6\\
PolyViT, $L_{adapt} = 0$ & 1 & \textbf{93M} & \cellcolor[rgb]{0.83, 1.0, 0.37} 83.1 & \cellcolor[rgb]{0.62, 1.0, 0.45} 91.2 & \cellcolor[rgb]{0.51, 1.0, 0.49} 99.0 & \cellcolor[rgb]{0.53, 1.0, 0.49} 95.0 & \cellcolor[rgb]{0.5, 1.0, 0.5} \textbf{96.7} & \cellcolor[rgb]{0.67, 1.0, 0.43} 77.5 & \cellcolor[rgb]{0.8, 1.0, 0.38} 33.2 & \cellcolor[rgb]{0.67, 1.0, 0.43} 32.3 & \cellcolor[rgb]{0.53, 1.0, 0.49} 50.6\\
PolyViT, $L_{adapt} = L/2$ & 1 & 178M & \cellcolor[rgb]{0.92, 1.0, 0.33} 82.8 & \cellcolor[rgb]{0.61, 1.0, 0.46} 91.5 & \cellcolor[rgb]{0.51, 1.0, 0.49} 99.0 & \cellcolor[rgb]{0.53, 1.0, 0.49} 95.0 & \cellcolor[rgb]{0.51, 1.0, 0.5} 96.6 & \cellcolor[rgb]{0.55, 1.0, 0.48} 79.4 & \cellcolor[rgb]{0.61, 1.0, 0.46} 35.3 & \cellcolor[rgb]{0.64, 1.0, 0.44} 33.1 & \cellcolor[rgb]{0.51, 1.0, 0.5} 51.5\\

\bottomrule

\end{tabular}
}
\vspace{-1.5\baselineskip}
\end{center}
\end{table*}

We set the training hyperparameters for these tasks (and those of the single-task baselines) using the values reported by~\cite{vit} for image tasks, \cite{vivit} for video tasks and \cite{mbt} for audio tasks (detailed in Appendix~\ref{sec:addexp}).
Note that the ``audio-only'' model of~\cite{mbt}, which we use as our baseline, is identical to AST~\cite{gong2021ast}, and we choose it since the authors have evaluated on more datasets.

We perform experiments with two standard transformer encoder configurations:
Base (number of layers, $L = 12$, hidden dimension $d = 768$, attention heads $h = 12$) and Large ($L = 24$, $d = 1024$, $h = 16$) following~\cite{bert, vit}.
As in~\cite{vit}, we initialize our PolyViT model and baselines with ViT pretrained on ImageNet-21K.
We refer to this initialized model as ViT-Im21K. %
For reproducibility, we will release code and models upon acceptance, and include exhaustive experimental details in Appendix~\ref{sec:addexp}.

\subsection{Selecting the best task sampling schedule for co-training}
\label{sec:experiments_schedule}

We begin by analyzing the effect of the different task sampling schedules listed in Sec.~\ref{sec:sch}.
We use the full, aforementioned 9-task set-up with
PolyViT-Base and all encoder layers shared ($L_{shared} = L = 12$, $L_{adapt} = 0$).

As shown in Tab.~\ref{tab:sch}, the ``Task-by-task'' schedule performs poorly, and only achieves decent performance on one task, %
 as it suffers from catastrophic forgetting~\cite{french1999catastrophic}.
The ``Accumulated'' sampling strategy requires using a single learning rate for all tasks (since the accumulated %
gradient over all tasks is used for performing a parameter update).
As we used a learning rate of 0.03, which is the learning rate used by the image tasks, and significantly lower than the learning rates for the video and audio tasks of the baselines (details in Appendix~\ref{sec:addexp}), this method only performs well on image datasets.
The ``Alternating'', ``Uniform'' and ``Weighted'' strategies perform the best, showing that task-specific learning rates, and switching between gradient-updates for different tasks is crucial for accuracy. %

In particular, the ``Weighted'' sampling method performs the best, achieving the highest accuracies on 8 of the 9 tasks (and second-highest on the remainder), motivating us to use it for all subsequent experiments.
Note that the ``Weighted'' strategy samples tasks with a lower number of training steps in their baseline training configurations less frequently.
In particular, the Pets task is only sampled for 500 iterations, out of the 417 000 total steps, or just 0.11\% of the SGD updates.
Nevertheless, it still achieves the highest accuracy on this task.
Another advantage of the ``Weighted'' strategy is that it performs the same number of steps per task as a single-task baseline.
Therefore, it uses the same computational resources during training as 9 separate, single-task baselines.
Our experiment also shows that if we do not have training hyperparameters for a new task, we can simply tune them separately in the single-task setting, and then reuse them for co-training.
This approach requires significantly less computation than tuning training hyperparameters directly in the co-training setup.

\begin{table*}[t]
\caption{Linear probing of PolyViT and single-task baselines. Similar to the protocol for evaluating self-supervised representation learning, we train only a linear classifier on top of a ``frozen'' transformer encoder.
Note how PolyViT co-trained on all tasks transfers well to all other datasets and modalities.
Models trained on audio do not transfer well to images and video, and vice versa.
All models are pretrained on ImageNet-21K, and then optionally finetuned on downstream datasets.
}
\label{tab:linev}
\begin{center}
\vspace{-2\baselineskip}
\scalebox{0.75}{
\begin{tabular}{lc cccccc ccc cc}
\\ \toprule
\multicolumn{2}{c}{}
&\multicolumn{6}{c}{Image}
&\multicolumn{3}{c}{Video}
&\multicolumn{2}{c}{Audio} \\
\cmidrule(r){3-8}\cmidrule(lr){9-11}\cmidrule(l){12-13}
\multicolumn{1}{l}{Model}
&\multicolumn{1}{c}{Finetuning}
&\multicolumn{1}{c}{\rotatebox{90}{C-ch101}}
&\multicolumn{1}{c}{\rotatebox{90}{SUN397}}
&\multicolumn{1}{c}{\rotatebox{90}{Dmlab}}
&\multicolumn{1}{c}{\rotatebox{90}{DTD}}
&\multicolumn{1}{c}{\rotatebox{90}{KITTI}}
&\multicolumn{1}{c}{\rotatebox{90}{PCAM}}
&\multicolumn{1}{c}{\rotatebox{90}{Epic K.}}
&\multicolumn{1}{c}{\rotatebox{90}{S-S v2}}
&\multicolumn{1}{c}{\rotatebox{90}{K600}}
&\multicolumn{1}{c}{\rotatebox{90}{MiT}}
&\multicolumn{1}{c}{\rotatebox{90}{K400}}
\\
\cmidrule(r){1-2} \cmidrule(r){3-8}\cmidrule(lr){9-11}\cmidrule(l){12-13}

ViT-Im21K pretrained & -- & \cellcolor[rgb]{0.53, 1.0, 0.49} 88.9 & \cellcolor[rgb]{0.56, 1.0, 0.48} 75.7 & \cellcolor[rgb]{0.8, 1.0, 0.38} 41.0 & \cellcolor[rgb]{0.51, 1.0, 0.5} 72.1 & \cellcolor[rgb]{1.0, 0.85, 0.41} 46.9 & \cellcolor[rgb]{1.0, 0.87, 0.4} 80.2 & \cellcolor[rgb]{1.0, 0.98, 0.32} 10.0 & \cellcolor[rgb]{0.95, 1.0, 0.32} 17.8 & \cellcolor[rgb]{0.69, 1.0, 0.42} 66.6 & \cellcolor[rgb]{1.0, 0.6, 0.6} 4.9 & \cellcolor[rgb]{1.0, 0.6, 0.6} 10.8\\
\midrule ViT & ImageNet-1K & \cellcolor[rgb]{0.5, 1.0, 0.5} \textbf{91.0} & \cellcolor[rgb]{0.51, 1.0, 0.5} 79.3 & \cellcolor[rgb]{0.52, 1.0, 0.49} 45.6 & \cellcolor[rgb]{0.52, 1.0, 0.49} 71.9 & \cellcolor[rgb]{0.71, 1.0, 0.42} 52.5 & \cellcolor[rgb]{1.0, 0.94, 0.34} 80.7 & \cellcolor[rgb]{0.88, 1.0, 0.35} 12.2 & \cellcolor[rgb]{0.92, 1.0, 0.33} 18.5 & \cellcolor[rgb]{0.67, 1.0, 0.43} 67.9 & \cellcolor[rgb]{1.0, 0.66, 0.56} 5.3 & \cellcolor[rgb]{1.0, 0.69, 0.54} 12.0\\
PolyViT & Image tasks & \cellcolor[rgb]{0.5, 1.0, 0.5} 90.7 & \cellcolor[rgb]{0.5, 1.0, 0.5} \textbf{80.0} & \cellcolor[rgb]{0.54, 1.0, 0.48} 45.2 & \cellcolor[rgb]{0.5, 1.0, 0.5} \textbf{72.5} & \cellcolor[rgb]{0.59, 1.0, 0.46} 53.8 & \cellcolor[rgb]{0.98, 1.0, 0.31} 81.2 & \cellcolor[rgb]{0.89, 1.0, 0.34} 12.1 & \cellcolor[rgb]{0.95, 1.0, 0.32} 17.9 & \cellcolor[rgb]{0.67, 1.0, 0.43} 67.9 & \cellcolor[rgb]{1.0, 0.66, 0.56} 5.3 & \cellcolor[rgb]{1.0, 0.68, 0.54} 11.9\\
\midrule ViViT & MiT & \cellcolor[rgb]{0.59, 1.0, 0.46} 85.2 & \cellcolor[rgb]{0.59, 1.0, 0.47} 73.8 & \cellcolor[rgb]{0.68, 1.0, 0.43} 43.0 & \cellcolor[rgb]{0.57, 1.0, 0.47} 69.9 & \cellcolor[rgb]{0.5, 1.0, 0.5} \textbf{54.9} & \cellcolor[rgb]{0.89, 1.0, 0.34} 81.7 & \cellcolor[rgb]{0.71, 1.0, 0.42} 14.9 & \cellcolor[rgb]{0.6, 1.0, 0.46} 26.3 & \cellcolor[rgb]{0.58, 1.0, 0.47} 74.2 & \cellcolor[rgb]{1.0, 0.63, 0.58} 5.1 & \cellcolor[rgb]{1.0, 0.68, 0.54} 11.9\\
PolyViT & Video tasks & \cellcolor[rgb]{0.53, 1.0, 0.49} 89.2 & \cellcolor[rgb]{0.53, 1.0, 0.49} 77.5 & \cellcolor[rgb]{0.5, 1.0, 0.5} \textbf{45.9} & \cellcolor[rgb]{0.54, 1.0, 0.49} 71.1 & \cellcolor[rgb]{0.62, 1.0, 0.45} 53.5 & \cellcolor[rgb]{0.5, 1.0, 0.5} \textbf{83.8} & \cellcolor[rgb]{0.55, 1.0, 0.48} 17.2 & \cellcolor[rgb]{0.53, 1.0, 0.49} 27.9 & \cellcolor[rgb]{0.5, 1.0, 0.5} \textbf{79.7} & \cellcolor[rgb]{1.0, 0.66, 0.56} 5.3 & \cellcolor[rgb]{1.0, 0.7, 0.52} 12.2\\
\midrule AST & VGGSound & \cellcolor[rgb]{1.0, 0.6, 0.6} 29.0 & \cellcolor[rgb]{1.0, 0.6, 0.6} 7.6 & \cellcolor[rgb]{1.0, 0.6, 0.6} 29.8 & \cellcolor[rgb]{1.0, 0.6, 0.6} 34.7 & \cellcolor[rgb]{1.0, 0.73, 0.5} 45.1 & \cellcolor[rgb]{1.0, 0.76, 0.48} 79.5 & \cellcolor[rgb]{1.0, 0.6, 0.6} 2.9 & \cellcolor[rgb]{1.0, 0.6, 0.6} 4.6 & \cellcolor[rgb]{1.0, 0.6, 0.6} 10.6 & \cellcolor[rgb]{0.61, 1.0, 0.46} 9.7 & \cellcolor[rgb]{0.53, 1.0, 0.49} 21.7\\
PolyViT & Audio tasks & \cellcolor[rgb]{1.0, 0.73, 0.51} 38.8 & \cellcolor[rgb]{1.0, 0.68, 0.54} 14.7 & \cellcolor[rgb]{1.0, 0.68, 0.54} 31.4 & \cellcolor[rgb]{1.0, 0.71, 0.51} 40.1 & \cellcolor[rgb]{1.0, 0.6, 0.6} 43.2 & \cellcolor[rgb]{1.0, 0.6, 0.6} 78.4 & \cellcolor[rgb]{1.0, 0.61, 0.6} 3.0 & \cellcolor[rgb]{1.0, 0.64, 0.57} 5.8 & \cellcolor[rgb]{1.0, 0.65, 0.57} 14.5 & \cellcolor[rgb]{0.5, 1.0, 0.5} \textbf{10.3} & \cellcolor[rgb]{0.5, 1.0, 0.5} \textbf{22.0}\\
\midrule PolyViT $L_{adapt} \! = \! 0$ & All & \cellcolor[rgb]{0.5, 1.0, 0.5} \textbf{91.0} & \cellcolor[rgb]{0.52, 1.0, 0.49} 78.2 & \cellcolor[rgb]{0.51, 1.0, 0.5} 45.8 & \cellcolor[rgb]{0.52, 1.0, 0.49} 71.8 & \cellcolor[rgb]{0.72, 1.0, 0.41} 52.3 & \cellcolor[rgb]{0.85, 1.0, 0.36} 81.9 & \cellcolor[rgb]{0.58, 1.0, 0.47} 16.8 & \cellcolor[rgb]{0.53, 1.0, 0.49} 27.9 & \cellcolor[rgb]{0.53, 1.0, 0.49} 77.8 & \cellcolor[rgb]{0.63, 1.0, 0.45} 9.6 & \cellcolor[rgb]{0.62, 1.0, 0.45} 20.6\\
PolyViT $L_{adapt} \! = \! L/2$ & All & \cellcolor[rgb]{0.5, 1.0, 0.5} 90.7 & \cellcolor[rgb]{0.53, 1.0, 0.49} 77.8 & \cellcolor[rgb]{0.55, 1.0, 0.48} 45.1 & \cellcolor[rgb]{0.51, 1.0, 0.5} 72.1 & \cellcolor[rgb]{0.71, 1.0, 0.42} 52.5 & \cellcolor[rgb]{0.78, 1.0, 0.39} 82.3 & \cellcolor[rgb]{0.5, 1.0, 0.5} \textbf{18.0} & \cellcolor[rgb]{0.5, 1.0, 0.5} \textbf{28.7} & \cellcolor[rgb]{0.5, 1.0, 0.5} 79.4 & \cellcolor[rgb]{0.57, 1.0, 0.47} 9.9 & \cellcolor[rgb]{0.58, 1.0, 0.47} 21.1\\
\bottomrule
\end{tabular}
} %
\vspace{-1.5\baselineskip}
\end{center}
\end{table*}

\subsection{Co-training with PolyViT} \label{sec:cotr}

Table~\ref{tab:cotr} presents approaches for training models to solve 9 different tasks across the image, video and audio modalities.
We consider two variants of PolyViT:
The first is PolyViT for a single modality, where we co-train three separate PolyViT models on all the tasks from either the image, video or audio modalities.
The second is the multi-modal PolyViT scenario where we co-train on all nine tasks across three modalities.
Here, we set $L_{adapt}$  to  0 and  $ L/2$ respectively to understand the effect of the number of modality-adaptor and shared layers.

We compare PolyViT to two baselines, which illustrate two alternatives to co-training.
One baseline is to train 9 separate single-task models for each dataset, either ViT, ViViT or AST depending on the modality.
This results in accuracies comparable to the state-of-the-art on the respective datasets, but also the largest number of total parameters.
The second baseline is to use a ViT model initialized on ImageNet-21K (ViT-Im21K) and to ``freeze'' the encoder of the network and train only the linear output heads (Eq.~\ref{eq:vit4},\ref{eq:theads}) for each task.
Positional embeddings, $\mathbf{p}$, and input embeddings, $\mathbf{E}$ are initialized following the methods used by ViT, ViViT or AST as described in Sec.~\ref{sec:vit}.
This baseline has the same number of parameters as PolyViT with $L_{adapt} = 0$.

Table~\ref{tab:cotr} shows that PolyViT trained on a single modality achieves the highest performance on 7 of the 9 datasets.
On the remaining two, the accuracy difference is negligible, as it is at most 0.3\%.
Moreover, the total number of parameters is 3 times less than the single-task baselines.
Single-modality co-training improves accuracy the most on the smaller datasets within the modality (Kinetics 400 in video, Mini-AudioSet for audio, and CIFAR-100 for images; full dataset details in Appendix~\ref{sec:addexp}).
This suggests that co-training acts as a regularizer, as noted by~\cite{caruana1997multitask}, that facilitates learning on smaller datasets where high-capacity models would otherwise overfit.

Multi-modal PolyViT (final two rows) achieves competitive performance whilst using substantially fewer parameters.
In particular, PolyViT with all transformer encoder layers shared between modalities ($L_{adapt} = 0$) is within 1.2\% of the single-task baselines across all datasets whilst using 8.3 times fewer parameters.
This model also comprehensively outperforms the ViT-Im21K Linear probe baseline which has the same number of parameters.
Sharing half the transformer layers between modalities ($L_{adapt} = L / 2$) increases model capacity, and the model improves upon the corresponding single-task baseline on 4 datasets, whilst being at most 0.5\% worse on the others.
The total number of parameters is still reduced by a factor of 4.3 compared to the single-task baselines.

Our results are consistent when using the Large model backbone as shown in Appendix \ref{sec:addexpcotr}.

\subsection{Evaluating learned representations with linear probes}
\label{sec:experiments_linearprobe}

We now evaluate the feature representations learned by PolyViT by simply appending and training only a new linear head (Eq.~\ref{eq:vit4},\ref{eq:theads}) for a new task.
This evaluation therefore follows the experimental setting commonly used in self-supervised learning to evaluate the quality of learned representations~\cite{chen2020simple, grill2020bootstrap}.
Note that the new task can come from any one of the three modalities that PolyViT is trained on, since the modality-adaptor layers (if present) are modality-specific rather than task-specific.

In particular, we evaluate on a number of new image, audio and video datasets as detailed in Appendix \ref{sec:addlp}.
For image classification, we include Caltech101, SUN397, DmLab, DTD, Kitti Distance and PatchCamelyon, which are datasets from the Visual Task Adaptation Benchmark~\cite{zhai2019large} not in our co-training set.
For video classification, we also include Epic Kitchens, Something-Something v2 and Kinetics 600.
Finally, for audio classification, we use the audio versions of Moments in Time and Kinetics 400.

We use PolyViT-Base and take linear probes of all the PolyViT models from Sec.~\ref{sec:cotr}, i.e. three single-modality models and two multi-modal models trained on all tasks with $L_{adapt} = 0$ and $L_{adapt} = L / 2$ respectively.
Our baseline models are those not performing co-training.
Namely, we use ViT trained only on ImageNet-21K (ViT-Im21K) as a baseline, followed by ViT, ViViT and AST initialized from ViT-Im21K and finetuned on ImageNet, Moments in Time and VGGSound respectively (since these are the largest datasets for each respective modality).

Table~\ref{tab:linev} shows how PolyViT trained on multiple modalities learns cross-modal feature representations that perform well on all 11 linear evaluation tasks across three different modalities (last two rows).
This holds even when all the layers of the PolyViT transformer layer are shared, and thus the total number of parameters is roughly equal to a single-task model.
PolyViT where the first half of the transformer encoder layers are modality-specific (final row), has more parameters and in general performs better.
Furthermore, for the Epic Kitchens (video), Something-Something v2 (video) and Caltech 101 (image) datasets, multi-modal PolyViT transfers better than single-modality baselines.
Table~\ref{tab:linev} thus demonstrates how co-training on multiple modalities facilitates learning powerful, transferable feature representations that can be used on multiple downstream tasks.

Models trained on only a single modality, as expected, do not in general learn feature representations that transfer well to other modalities.
In particular, models trained on audio tasks do not transfer at all to images and videos, and vice versa.
Models trained on video, however, still perform well on images, with video-trained models performing the best on the DmLab, PCAM and KITTI-Distance datasets.
We believe this is due to the commonalities between the image and video modalities.
Observe that in the majority of cases, single-modality PolyViT models perform better on linear probing than the corresponding single-task baselines, especially for video and audio.

\subsection{State-of-the-art performance with single-modality co-training} \label{sec:ls1mod}
\begin{table}[t]
\centering
\caption{
Comparison to MBT~\cite{mbt}, the current published state-of-the-art using the same protocols.
The second and third rows show that MBT, when first trained on AudioSet and then finetuned on VGGSound, and vice-versa, does not perform as well as PolyViT, showing that the regularizing benefits of co-training are not simply because the co-trained model has access to more data.
}
\label{tab:exp_audio_sota}
\scalebox{0.68}{
\begin{tabular}{lcc c cc}
\toprule
      &          &          & \multicolumn{2}{c}{VGGSound} & AudioSet\\ 
Model & \#Models & \#Params & Top 1          & Top 5 & mAP  \\
\cmidrule(r){1-1} \cmidrule(lr){2-3} \cmidrule(lr){4-5} \cmidrule(l){6-6}
MBT (audio-only) & 2 & 172M  & 52.3 & 78.1 & 44.3 \\
MBT: AS500k $\to$ VGGSound & 1 & 87M  & 54.4 & 81.4 & 34.2 \\
MBT: VGGSound $\to$ AS500k & 1 & 87M  & 22.1 & 43.5 & 44.4 \\
\midrule 
PolyViT & 1 & \textbf{87M}  & \textbf{55.1\iffalse55.07\fi} & \textbf{80.4\iffalse80.40\fi} & \textbf{44.5\iffalse44.52\fi}\\
\bottomrule
\end{tabular}
}
\end{table}

Motivated by the performance of single-modality co-training in Tab.~\ref{tab:cotr}, we perform larger-scale co-training experiments with this method on audio and video classification.
Tables~\ref{tab:exp_audio_sota} and \ref{tab:vidls} show that we achieve state-of-the-art results in both of these domains whilst using also significantly fewer parameters.

On audio classification, we compare to the current state-of-the-art using audio information only, MBT~\cite{mbt}, using the same Base backbone, training on the balanced AS-500k subset, and other experimental settings as the authors~\cite{mbt}.
As shown in Tab.~\ref{tab:exp_audio_sota}, we surpass the state-of-the-art on both datasets (AudioSet and VGGSound), whilst using about half the total parameters.
We observe larger improvements (2.8\%) on VGGSound, the smaller dataset.
This is line with our findings from Sec.~\ref{sec:cotr} and shows that co-training has a regularizing effect that reduces overfitting and improves performance the most on smaller datasets.
The second and third rows of Tab.~\ref{tab:exp_audio_sota} also show that training MBT on AudioSet and then finetuning on VGGSound, or vice versa, produces worse results than our co-training method.
This shows that the regularization benefits of co-training are not solely from having access to more data than single-task baselines.
As expected, finetuning MBT on the target dataset causes accuracy to degrade on the original dataset.

\begin{table}[t]
\centering
\caption{
Comparison to ViVIT~\cite{vivit}, the current published state-of-the-art, using the same experimental settings as~\cite{vivit}.
K400 and K600 denote Kinetics-400 and -600 respectively. MiT denotes the Moments in Time dataset.
}
\label{tab:vidls}
\scalebox{0.68}{
\begin{tabular}{lcc cc cc cc}
\toprule
      &          &          & \multicolumn{2}{c}{K400} & \multicolumn{2}{c}{K600} & \multicolumn{2}{c}{MiT} \\ 
Model & \#Models & \#Params & Top 1           & Top 5          & Top 1           & Top 5          & Top 1            & Top 5            \\
\cmidrule(r){1-1} \cmidrule(lr){2-3} \cmidrule(lr){4-5} \cmidrule(lr){6-7} \cmidrule(lr){8-9}
ViViT & 3 & 913M & 80.6 & 94.7 & 82.5 & \textbf{95.6} & 38.0 & 64.9 \\
PolyViT & 1 & \textbf{308M} & \textbf{82.4\iffalse82.38\fi} & \textbf{95.0\iffalse94.97\fi} & \textbf{82.9\iffalse82.85\fi} & 95.5\iffalse95.50\fi & \textbf{38.6\iffalse38.57\fi} & \textbf{65.5\iffalse65.45\fi} \\
\bottomrule
\end{tabular}
}
\end{table}

For video classification, we co-train PolyViT-Large with a smaller tubelet size (and hence greater number of tokens) of $2 \times 16 \times 16$ on Kinetics-400, -600 and Moments in Time.
We compare to ViViT~\cite{vivit} which is the current published, state-of-the-art, and uses the same initialization, backbone and number of tokens.
As shown in Tab.~\ref{tab:vidls}, we surpass the state-of-the-art on all three datasets.
Once again, the largest improvement of 1.8\% is on Kinetics 400, which is also the smallest dataset, as co-training has a regularizing effect.
Moreover, by co-training on three datasets, we reduce the total number of parameters required by almost three compared to separately trained ViViT models. 
Appendix~\ref{sec:addls} compares our models to other previous works on these audio and video datasets.

\section{Conclusion}

By co-training PolyViT on a single modality, we have achieved state-of-the-art results on three video and two audio datasets, while reducing the total number of parameters linearly compared to single-task models.
PolyViT co-trained on multiple modalities is even more parameter-efficient, still competitive with the state-of-the-art, and learns feature representations that generalize across multiple modalities.
This enables us to learn new tasks by simply learning an additional output head.
Co-training is simple and practical, as we do not need to tune hyperparameters on the joint space of all datasets, but can simply re-use training hyperparameters from single-task models.
Moreover, we can achieve accuracy improvements from training for the same number of total steps.

\paragraph{Limitations and future work}
Current limitations of our method are that we do not co-train on large-scale upstream datasets such as ImageNet-21K~\cite{imagenet} and C4~\cite{raffel2019exploring}.
We aim to explore this, and co-training with the text modality, in future work.
As aforementioned, our model, although versatile, does not improve inference time as it still processes a single task at a time.
We also do not currently fuse multiple modalities together (ie video and audio) to make a better prediction, and aim to do so in future.

\section*{Acknowledgements}
We would like to thank Xiaohua Zhai, Neil Houlsby and Cordelia Schmid for the discussions along the way, and their comments and feedback on the paper. We also thank Arsha Nagrani and Chen Sun for sharing MBT code with us. Finally, we thank the Google Brain team at large for providing a supportive research environment.

{\small
\bibliographystyle{ieee_fullname}
\bibliography{references}
}

\newpage
\onecolumn
\appendix

\section*{Appendix}

Appendix~\ref{sec:addexp} contains additional details about our experimental settings, providing more information to Section 5.1 of the main paper.
Appendix~\ref{sec:addsch} provides more details for the experiments in Section 5.2 of the main paper.
Appendix~\ref{sec:addexpcotr} shows further experimental details and results corresponding to Section 5.3 of the main paper.
Appendix~\ref{sec:addlp} provides additional details about the experiments in Section 5.4 of the paper.
Finally. Appendix~\ref{sec:addls} provides additional experimental details and results corresponding to Section 5.5 of the main paper. 

\subsection*{Broader Impact}
    Our work presents a method for performing image-, audio- and video-classification with a single parameter-efficient model.
    Classification of perceptual data (images, audio and video) is a general technology with a wide range of potential applications.
    While we are unaware of all potential applications, it is important to be aware that each application has its own merits and societal implications depending on the intentions of the individuals building and using the system.
    We also note that training datasets contain biases that may render models trained on them unsuitable for certain applications.
    It is possible that people use classification models (intentionally or not) to make decisions that impact different groups in society differently.

\section{Experimental set-up: additional details} \label{sec:addexp}

\paragraph{Task details and input dimensions.} See Tables \ref{tab:ds} and \ref{tab:mod}. For each task, the number of linear warmup steps is set as reported in \cite{vit,vivit,mbt}. When co-training, we simply use the sum of all warmup steps for each co-trained task. We use a single momentum state when cotraining, i.e. we don't maintain separate momentum states for each task or modality. Similar to \cite{vit}, we select the best learning rate on a set $\{ 0.03, 0.1, 0.3 \}$ using the validation score. For video and audio datasets, we reuse learning rates reported in \cite{vivit} and \cite{mbt} respectively. As in \cite{vivit,mbt}, we use zero initialization for output head kernels $\mathbf{W}_{out}$. For image datasets, on a single-task evaluation, we find that LeCun normal $\mathbf{W}_{out}$ initializer \cite{lecunn} works best. For the ``ViT-Im21K linear probe'' baseline, we use the same training procedure as for single-task baselines, with the difference that 1) only the head parameters are updated and 2) on image tasks, we run separate learning rate grid searches on the set $\{ 0.03, 0.1, 0.3 \}$.

\paragraph{Train, validation and test splits.} Similarly to \cite{vit}, we take 2\% of CIFAR 10/100 train sets for validation, 10\% of Pets train set for validation and 1\% of ImageNet-1k train set for validation. We use standard test sets for these datasets. For RESISC45, we use 20\% of the train set for validation and 20\% for testing. We use standard train, validation and test sets for video and audio tasks.

\paragraph{Augmentation and regularization.} We don't use augmentation for image tasks. We do video and audio preprocessing and augmentation as done in \cite{vivit,mbt} respectively. For audio tasks, as in \cite{mbt}, we use Mixup \cite{mixup} with $\alpha = 0.3$ and stochastic depth regularization \cite{stdepth} with $p = 0.3$. Stochastic depth is applied along both audio adaptor and shared layers.

\begin{table*}[h]
\centering
\caption{Experimental set-up: tasks and their properties. For image tasks, the indicated learning rates are obtained by a grid search over $\{ 0.03, 0.1, 0.3 \}$ on single-task baselines using the validation set accuracy. These values are used for single-task baselines and for PolyViT variants.
}
\label{tab:ds}
\scalebox{0.9}{
\begin{tabular}{ccccccccc}
\toprule
\multicolumn{1}{c}{ Dataset}
&\multicolumn{1}{c}{ \makecell{Abbre-\\viation}}
&\multicolumn{1}{c}{ \makecell{Moda-\\lity}}
&\multicolumn{1}{c}{ \makecell{Clas-\\ses}}
&\multicolumn{1}{c}{ \makecell{Train \\ size}}
&\multicolumn{1}{c}{ \makecell{Train \\ steps}}
&\multicolumn{1}{c}{ \makecell{Learning \\ rate}}
&\multicolumn{1}{c}{ \makecell{Warmup \\ steps}}
&\multicolumn{1}{c}{ \makecell{$\mathbf{W}_{out}$ \\ init}} \\
\midrule
CIFAR 100 & C100 & Image & 100 & 50.0K & 10K & 0.03 & 500 & \makecell{LeCun \\ normal} \\
CIFAR 10 & C10 & Image & 10 & 50.0K & 10K & 0.03 & 500 & \makecell{LeCun \\ normal} \\
\makecell{Oxford-IIIT \\ Pets} & Pets & Image & 37 & 3.68K & 500 & 0.03 & 100 & \makecell{LeCun \\ normal} \\
RESISC45 & R45 & Image & 45 & 31.5K & 2.5K & 0.1 & 200 & \makecell{LeCun \\ normal} \\
ImageNet-1k & Im1K & Image & 1000 & 1.28M & 20K & 0.03 & 500 & \makecell{LeCun \\ normal} \\
\midrule
Kinetics 400 & K400 & Video & 400 & 215K & \makecell{100.7K \\ (30 epochs)} & 0.1 & 2.5 epochs & Zeros \\
\makecell{Moments in \\ Time} & MiT & Video & 339 & 791K & \makecell{123.6K \\ (10 epochs)} & 0.25 & 2.5 epochs & Zeros \\
\midrule Mini-Audioset & MiniAS & Audio & 527 & 20.4K & \makecell{15.9K \\ (50 epochs)} & 0.5 & 2.5 epochs & Zeros \\
VGGSound & VGG & Audio & 309 & 172K & \makecell{135K \\ (50 epochs)} & 0.5 & 2.5 epochs & Zeros \\
\bottomrule
\end{tabular}
}
\end{table*}

\begin{table*}[t]
\caption{Input dimensions for different modalities. Sequence length is computed as $1 + [(T / t) \times ] (H / h) \times (W / w)$ (one class token and patch tokens). Note that for shared transformer layers, we reuse the same parameters for sequences of different lengths.
}
\label{tab:mod}
\begin{center}
\begin{tabular}{ccccc}
\toprule
\multicolumn{1}{c}{Modality}
&\multicolumn{1}{c}{\makecell{Input size, \\ $[T \times] H \times W$}}
&\multicolumn{1}{c}{\makecell{Patch size, \\ $[t \times] h \times w$}}
&\multicolumn{1}{c}{\makecell{Sequence \\ length}}
&\multicolumn{1}{c}{\makecell{Batch size}} \\
\cmidrule(r){1-5} Image (pretraining) & $224 \times 224$ & $16 \times 16$ & 197  & 4096 \\
Image & $384 \times 384$ & $16 \times 16$ & 577 & 512 \\
Video & $32 \times 224 \times 224$ & $4 \times 16 \times 16$ & 1569 & 64 \\
Audio (spectrogram) & $800 \times 128$ & $16 \times 16$ & 401 & 64 \\
\bottomrule
\end{tabular}
\end{center}
\end{table*}

\section{Selecting the best task sampling schedule: additional experimental details} \label{sec:addsch}

For the accumulating schedule, we set learning rate to the smallest value across tasks ($0.03$). We draw a random task order for the Task-by-task schedule, which is as follows: C100 $\to$ MiT $\to$ K400 $\to$ MiniAS $\to$ VGG $\to$ Pets $\to$ C10 $\to$ Im1K $\to$ R45.

\section{Co-training with PolyViT: additional experimental details and results} \label{sec:addexpcotr}

\paragraph{Evaluation on video and audio tasks.} To get test performance on video and audio tasks, we perform multiple-crop evaluation as described in \cite{vivit, mbt} for videos and audio respectively.

\paragraph{Results for the Large configuration.} See Table \ref{tab:cotr-l}. Since \cite{mbt} don't report results on a Large configuration, for audio tasks we do an additional hyperparameter tuning for single-task baselines on validation sets. As a result, we use Mixup $\alpha = 0.5, 0.7$ for MiniAS and VGGSound respectively. Also, we use $30$ epochs for MiniAS instead of 50 for the Base model. In addition, we run separate learning rate grid searches for all image tasks, separately for single-task baselines and ViT-Im21K linear probes. We apply all mentioned hyperparameter changes, obtained for the single-task baselines, to all PolyViT runs. In all other aspects, Large set-up is the same as Base.

\begin{table*}[th]
\caption{
Co-training with PolyViT, Large model configuration. Test accuracy (\%) and mAP (for MiniAS, \%) are reported. As indicated by the ``\# models'' column, some rows correspond to multiple models, then the total number of parameters is computed across all models.
}
\label{tab:cotr-l}
\begin{center}
\vspace{-\baselineskip}
\scalebox{0.75}{
    \begin{tabular}{l cc ccccc cc cccc}
    \toprule
             &  & & \multicolumn{5}{c}{Image}      & \multicolumn{2}{c}{Video} & \multicolumn{2}{c}{Audio} \\
             \cmidrule(lr){4-8} \cmidrule(lr){9-10} \cmidrule(l){11-12}
    Model & \#Models & \#Params &  C100 & C10 & Pets & R45 & Im1K & K400         & MiT        & MiniAS        & VGG \\
    \cmidrule(r){1-1} \cmidrule(lr){2-3} \cmidrule(lr){4-8} \cmidrule(lr){9-10} \cmidrule(l){11-12}

ViT-Im21k Linear probe & 1 & \textbf{312M} & \cellcolor[rgb]{1.0, 0.6, 0.6} 84.4 & \cellcolor[rgb]{1.0, 0.6, 0.6} 95.6 & \cellcolor[rgb]{1.0, 0.6, 0.6} 91.8 & \cellcolor[rgb]{1.0, 0.6, 0.6} 89.2 & \cellcolor[rgb]{1.0, 0.6, 0.6} 82.6 & \cellcolor[rgb]{1.0, 0.6, 0.6} 67.7 & \cellcolor[rgb]{1.0, 0.6, 0.6} 26.8 & \cellcolor[rgb]{1.0, 0.6, 0.6} 12.8 & \cellcolor[rgb]{1.0, 0.6, 0.6} 19.1\\
Single-task baseline & 9 & 3033M & \cellcolor[rgb]{0.56, 1.0, 0.47} 93.3 & \cellcolor[rgb]{0.55, 1.0, 0.48} 99.2 & \cellcolor[rgb]{0.69, 1.0, 0.42} 94.8 & \cellcolor[rgb]{0.5, 1.0, 0.5} \textbf{97.3} & \cellcolor[rgb]{0.5, 1.0, 0.5} \textbf{85.1} & \cellcolor[rgb]{0.61, 1.0, 0.46} 79.6 & \cellcolor[rgb]{0.64, 1.0, 0.44} 37.1 & \cellcolor[rgb]{0.81, 1.0, 0.37} 30.0 & \cellcolor[rgb]{0.5, 1.0, 0.5} \textbf{51.8}\\
\cmidrule(lr){1-12} PolyViT, 1 modality & 3 & 917M & \cellcolor[rgb]{0.5, 1.0, 0.5} \textbf{93.9} & \cellcolor[rgb]{0.5, 1.0, 0.5} \textbf{99.4} & \cellcolor[rgb]{0.5, 1.0, 0.5} \textbf{95.5} & \cellcolor[rgb]{0.55, 1.0, 0.48} 96.9 & \cellcolor[rgb]{0.5, 1.0, 0.5} \textbf{85.1} & \cellcolor[rgb]{0.53, 1.0, 0.49} 80.6 & \cellcolor[rgb]{0.5, 1.0, 0.5} \textbf{38.8} & \cellcolor[rgb]{0.5, 1.0, 0.5} \textbf{37.9} & \cellcolor[rgb]{0.53, 1.0, 0.49} 50.7\\
PolyViT, $L_{adapt} = 0$ & 1 & \textbf{312M} & \cellcolor[rgb]{0.76, 1.0, 0.39} 91.4 & \cellcolor[rgb]{0.61, 1.0, 0.46} 99.0 & \cellcolor[rgb]{0.72, 1.0, 0.41} 94.7 & \cellcolor[rgb]{0.56, 1.0, 0.48} 96.8 & \cellcolor[rgb]{1.0, 0.6, 0.6} 82.6 & \cellcolor[rgb]{0.66, 1.0, 0.44} 78.9 & \cellcolor[rgb]{0.75, 1.0, 0.4} 35.8 & \cellcolor[rgb]{0.68, 1.0, 0.43} 33.3 & \cellcolor[rgb]{0.56, 1.0, 0.48} 49.9\\
PolyViT, $L_{adapt} = L/2$ & 1 & 615M & \cellcolor[rgb]{0.79, 1.0, 0.38} 91.1 & \cellcolor[rgb]{0.58, 1.0, 0.47} 99.1 & \cellcolor[rgb]{0.64, 1.0, 0.45} 95.0 & \cellcolor[rgb]{0.54, 1.0, 0.49} 97.0 & \cellcolor[rgb]{1.0, 0.66, 0.55} 82.8 & \cellcolor[rgb]{0.5, 1.0, 0.5} \textbf{81.0} & \cellcolor[rgb]{0.59, 1.0, 0.46} 37.7 & \cellcolor[rgb]{0.65, 1.0, 0.44} 34.1 & \cellcolor[rgb]{0.54, 1.0, 0.48} 50.4 \\

\bottomrule

\end{tabular}
}
\vspace{-0.5cm}
\end{center}
\end{table*}

\section{Linear probes: additional experimental details} \label{sec:addlp}

\paragraph{Task details.} See Table \ref{tab:linds}. For linear probes, we use the same input dimensions as reported in Table \ref{tab:mod}. For image tasks, we reuse the number of train and warmup steps from the RESISC45 task (Table \ref{tab:ds}). For video and audio tasks, we used hyperparameters reported in \cite{vivit} and \cite{mbt} respectively, with the difference that we only optimize output head parameters during training. As for the co-training setup, we use multiple-crop evaluation on video and audio tasks.

\paragraph{Train, validation and test splits.} For image tasks, we use 2\% of the train set as a validation set and standard test sets. We use standard train, validation and test sets for video and audio tasks.

\paragraph{Converting patch and positional embeddings for cross-modal probes.} In order to take linear probes of image-only models (ViT and PolyViT trained on images) on audio tasks (and vice versa), we leave patch embeddings as they are and 2D-interpolate positional embeddings to the correct resolution. When taking linear probes of video-only models on image or audio tasks, in order to obtain $16 \times 16$ patch embeddings, we take a sum along the first (frame) axis of 3D video patch embeddings of shape $4 \times 16 \times 16$. In order to adapt positional embeddings, we take a mean value of positional embeddings for each frame, and then 2D-interpolate the result to the correct resolution. When taking linear probes of image- or audio-only
models on video tasks, we repeat 2D patch embeddings along the frame axis in order to obtain 3D patch embeddings. We also 2D-interpolate positional embeddings to the frame resolution and repeat them for each frame.

\paragraph{Augmentation and regularization.} We don't use augmentation for image tasks. We do video and audio preprocessing and augmentation as done in \cite{vivit,mbt} respectively. As in \cite{vivit}, we use Mixup \cite{mixup} with $\alpha = 0.3$ for the S-S v2 task.

\begin{table*}[h]
\caption{Tasks used for linear probes. Indicated learning rate grid search is done for all models using validation set performance.}
\label{tab:linds}
\begin{center}
\scalebox{0.9}{
\begin{tabular}{ccccccc}
\toprule
\multicolumn{1}{c}{\makecell{Dataset \\ (task)}}
&\multicolumn{1}{c}{\makecell{Abbre-\\viation}}
&\multicolumn{1}{c}{\makecell{Moda-\\lity}}
&\multicolumn{1}{c}{\makecell{Train \\ steps}}
&\multicolumn{1}{c}{\makecell{Learning \\ rate}}
&\multicolumn{1}{c}{\makecell{Warmup \\ steps}}
&\multicolumn{1}{c}{\makecell{$\mathbf{W}_{out}$ \\ init}} \\
\midrule
Caltech101 & C-ch101 & Image & 2.5K & \makecell{Grid search, \\ $\{ 0.03, 0.1, 0.3 \}$} & 200 & \makecell{LeCun \\ normal} \\
SUN397 & SUN397 & Image & 2.5K & \makecell{Grid search, \\ $\{ 0.03, 0.1, 0.3 \}$} & 200 & \makecell{LeCun \\ normal} \\
Dmlab & Dmlab & Image & 2.5K & \makecell{Grid search, \\ $\{ 0.03, 0.1, 0.3 \}$} & 200 & \makecell{LeCun \\ normal} \\
DTD & DTD & Image & 2.5K & \makecell{Grid search, \\ $\{ 0.03, 0.1, 0.3 \}$} & 200 & \makecell{LeCun \\ normal} \\
KITTI Distance & KITTI & Image & 2.5K & \makecell{Grid search, \\ $\{ 0.03, 0.1, 0.3 \}$} & 200 & \makecell{LeCun \\ normal} \\
PatchCamelyon & PCAM & Image & 2.5K & \makecell{Grid search, \\ $\{ 0.03, 0.1, 0.3 \}$} & 200 & \makecell{LeCun \\ normal} \\
\midrule
Epic Kitchens & Epic K. & Video & 30 epochs & 0.5 & 2.5 epochs & Zeros \\
Something-Something v2 & S-S v2 & Video & 35 epochs & 0.4 & 2.5 epochs & Zeros \\
Kinetics 600 & K600 & Video & 30 epochs & 0.1 & 2.5 epochs & Zeros \\
\midrule
Moments in Time (audio) & MiT-A & Audio & 10 epochs & 0.5 & 2.5 epochs & Zeros \\
Kinetics 400 (audio) & K400-A & Audio & 30 epochs & 0.5 & 2.5 epochs & Zeros \\
\bottomrule
\end{tabular}
}
\end{center}
\end{table*}

\section{State-of-the-art performance on one modality: additional experimental details and results} \label{sec:addls}

\paragraph{Additional results on video datasets}
Table~\ref{tab:vidls2} contains an extended comparison to ViViT~\cite{vivit}, compared to Table 5 of the main paper.
The second row (``ViVIT: MiT $\to$ K600 $\to$ K400'') shows our results when we train a ViViT model, by first finetuning an ImageNet-21K initialized model on Moments in Time, then Kinetics 600, and then finally Kinetics 400.
This model has seen the same amount of training data as PolyViT, but performs worse on Kinetics 400 than PolyViT (PolyViT achieves 82.4, and ViViT achieves 81.3).
This ViViT model, does however, still outperform a ViViT model finetuned solely on Kinetics 400 from ImageNet-21K initialization (first row).
This result, like Table 4 of the main paper for audio, shows that the benefits of co-training are not only because the co-trained PolyViT model has access to more data.

For our additional baseline, (``ViViT: MiT $\to$ K600 $\to$ K400''), we retain the output linear head for each class.
Consequently, the accuracy for MiT and Kinetics 600 degrades as the model is trained on Kinetics 400.
Note that Kinetics 600 is a superset of Kinetics 400, which is why the overall accuracy drop on Kinetics 600 is low.
Furthermore, note that the goal of this paper is not to consider the ``continual learning''~\cite{kirkpatrick2017overcoming} problem, which aims to train a model on a new dataset, without losing performance on previous datasets the model was trained on.

\paragraph{Comparison with other methods from the literature.}
Tables \ref{tab:as_sota}, \ref{tab:vgg_sota}, \ref{tab:sota_kinetics400}, \ref{tab:sota_kinetics600}, \ref{tab:sota_mit} are extended versions of Tables 4 and 5 from the main paper which include results for other methods from the literature for audio- and video-classification.

\paragraph{Detailed experimental settings}
For the PolyViT experiment on the video modality, we reuse hyperparameters reported in \cite{vivit} for Kinetics 400/600 and Moments in Time. See Table \ref{tab:videols} for the dataset details and exact hyperparameters used during the experiment. These hyperparameters coincide with those reported in Table \ref{tab:ds} for Kinetics 400 and Moments in Time and in Table \ref{tab:linds} for Kinetics 600. The only difference is that we use a more granular 3D patch size ($2 \times 16 \times 16$) and Large model configuration.

For the audio experiment, similarly, we reuse all hyperparameters reported in \cite{mbt} for AS-500k and VGGSound experiments (audio-only). See Table \ref{tab:audiols} for the dataset details and exact hyperparameters used for the experiment. These hyperparameters almost coincide with those reported in Table \ref{tab:linds} with a change MiniAS $\to$ AS-500k. The only exception is that we use 30 epochs and Mixup $\alpha = 0.5$ for AS-500k.

\begin{table}[t]
\centering
\caption{
Extended comparison to ViViT~\cite{vivit}.
The second row shows a ViVIT model, initialized from ImageNet-21K, and then finetuned on Moments in Time, Kinetics 600 and then Kinetics 400.
This model has seen the same amount of training data as PolyViT, yet does not perform as well as PolyViT, showing that the improvements from co-training are not solely because PolyViT has access to more training data.
}
\label{tab:vidls2}
\begin{tabular}{lcc cc cc cc}
\toprule
      &          &          & \multicolumn{2}{c}{K400} & \multicolumn{2}{c}{K600} & \multicolumn{2}{c}{MiT} \\ 
Model & \#Models & \#Params & Top 1           & Top 5          & Top 1           & Top 5          & Top 1            & Top 5            \\
\cmidrule(r){1-1} \cmidrule(lr){2-3} \cmidrule(lr){4-5} \cmidrule(lr){6-7} \cmidrule(lr){8-9}
ViViT & 3 & 913M & 80.6 & 94.7 & 82.5 & \textbf{95.6} & 38.0 & 64.9 \\
ViViT: MiT $\to$ K600 $\to$ K400 & 1 & \textbf{308M} & 81.3 & 94.5 & 78.8 & 94.0 & 27.3 & 52.0 \\
\midrule
PolyViT & 1 & \textbf{308M} & \textbf{82.4\iffalse82.38\fi} & \textbf{95.0\iffalse94.97\fi} & \textbf{82.9\iffalse82.85\fi} & 95.5\iffalse95.50\fi & \textbf{38.6}\iffalse38.57\fi & \textbf{65.5}\iffalse65.45\fi \\
\bottomrule
\end{tabular}
\end{table}

\begin{table}[t]
\centering
\caption{
AudioSet (audio only). Comparison with other methods from the literature.
}
\label{tab:as_sota}
\begin{tabular}{lcc}
\toprule
Model & Training set & mAP  \\
\cmidrule(r){1-3}
GBlend \cite{gblend} & MiniAS & 29.1 \\
GBlend \cite{gblend} & Full AS (2M) & 32.4 \\
Attn Audio-Visual \cite{fayek2020large} & Full AS (2M) & 38.4 \\
Perceiver \cite{perceiver} & Full AS (2M) & 38.4 \\
MBT \cite{mbt} & AS-500k & 44.3 \\
\midrule 
PolyViT & AS-500k & \textbf{44.5\iffalse44.52\fi}\\
\bottomrule
\end{tabular}
\end{table}

\begin{table}[t]
\centering
\caption{
VGGSound (audio only). Comparison with other methods from the literature.
}
\label{tab:vgg_sota}
\begin{tabular}{lcc}
\toprule
Model & Top 1 & Top 5  \\
\cmidrule(r){1-3}
Chen et al. \cite{vggsound} & 48.8 & 76.5 \\
AudioSlowFast \cite{slowfast} & 50.1 & 77.9 \\ 
MBT \cite{mbt} & 52.3 & 78.1 \\
\midrule 
PolyViT & \textbf{55.1} & \textbf{80.4} \\
\bottomrule
\end{tabular}
\end{table}

\begin{table}[tb]
		\centering
		\caption{Kinetics 400. Comparison with other methods from the literature.}
			\begin{tabular}{lcccc}
				\toprule
				Model 																			 & Top 1                & Top 5 \\
				\midrule
				blVNet~\cite{fan_blvnet_neurips_2019}							  & 73.5 				  & 91.2 \\ 
				STM~\cite{jiang_stm_iccv_2019}										& 73.7 					& 91.6\\
				TEA~\cite{li_tea_cvpr_2020}												& 76.1					& 92.5 \\  %
				TSM-ResNeXt-101~\cite{lin_tsm_cvpr_2019}						  & 76.3				 & -- \\
				I3D NL~\cite{wang_cvpr_2018}										 & 77.7                  & 93.3         		 \\  %
				CorrNet-101~\cite{wang_corrnet_cvpr_2020}					  & 79.2				 & --			 		\\  %
				ip-CSN-152~\cite{tran_iccv_2019}									&  79.2					& 93.8				\\  %
				LGD-3D R101~\cite{qiu_lgd_cvpr_2019}							&  79.4				    & 94.4			 		\\  
				SlowFast R101-NL~\cite{feichtenhofer_iccv_2019}       		&  79.8                 &  93.9                  \\  %
				X3D-XXL~\cite{feichtenhofer_cvpr_2020}      					&  80.4					&  94.6			  		\\
				TimeSformer-L~\cite{bertasius_arxiv_2021}					  	& 80.7				& 94.7					\\
				ViViT-L/16x2 (ImageNet-21K) \cite{vivit} 														  	 &  80.6 				& 94.7 \\  %
				MViT-B \cite{mvit} & 81.2 & 95.1 \\
				Mformer-HR \cite{mformer} & 81.1 & \textbf{95.2} \\
				\midrule
				PolyViT-L/16x2 & \textbf{82.4} & 95.0 \\
				\bottomrule
			\end{tabular}
		\label{tab:sota_kinetics400}
\end{table}

\begin{table}[tb]
		\centering
		\caption{Kinetics 600. Comparison with other methods from the literature.}
			\begin{tabular}{lcccc}
				\toprule
				Model 																			 & Top 1                & Top 5 \\
				\midrule
				AttentionNAS~\cite{wang_nas_eccv_2020}						  &  79.8				   & 94.4  \\ %
	  			LGD-3D R101~\cite{qiu_lgd_cvpr_2019}							&  81.5				    & 95.6	\\ %
	  			SlowFast R101-NL~\cite{feichtenhofer_iccv_2019}       		&  81.8                     &  95.1   \\ %
	  			X3D-XL~\cite{feichtenhofer_cvpr_2020}      						 &  81.9					&  95.5		\\ %
	  			TimeSformer-L~\cite{bertasius_arxiv_2021}					  & 82.2				& \textbf{95.6}		\\ %
	  			ViViT-L/16x2 (ImageNet-21K) \cite{vivit} & 82.5 & \textbf{95.6} \\
	  			Mformer-HR \cite{mformer} & 82.7 & 96.1 \\
	  			MViT-B-24 \cite{mformer} & \textbf{83.8} & 94.7 \\
				\midrule
				PolyViT-L/16x2 & 82.9 & 95.5 \\
				\bottomrule
			\end{tabular}
		\label{tab:sota_kinetics600}
\end{table}

\begin{table}[tb]
		\centering
		\caption{Moments in Time. Comparison with other methods from the literature.}
			\begin{tabular}{lcccc}
				\toprule
				Model 																			 & Top 1                & Top 5 \\
				\midrule
				TSN~\cite{wang_tsn_eccv_2016}				& 25.3		 &  50.1	\\
  				TRN~\cite{zhou_trn_eccv_2018}				 & 28.3		 &  53.4	 \\
  				I3D~\cite{carreira_cvpr_2017}					& 29.5		&  56.1		\\
  				blVNet~\cite{fan_blvnet_neurips_2019}	 &  31.4	 &  59.3 	 \\
  				AssembleNet-101~\cite{ryoo_iclr_2020} 	&  34.3     &  62.7     \\
	  			ViViT-L/16x2 (ImageNet-21K) \cite{vivit} & 38.0 & 64.9 \\
				\midrule
				PolyViT-L/16x2 & \textbf{38.6} & \textbf{65.5} \\
				\bottomrule
			\end{tabular}
		\label{tab:sota_mit}
\end{table}

\begin{table*}[h]
\centering
\caption{Set-up for the co-training on videos. Train steps and warmup steps are summed to get the number of train and warmup steps during co-training as we use the ``Weighted'' task sampling method.
}
\label{tab:videols}
\scalebox{0.9}{
\begin{tabular}{ccccccccc}
\toprule
\multicolumn{1}{c}{ Dataset}
&\multicolumn{1}{c}{ \makecell{Moda-\\lity}}
&\multicolumn{1}{c}{ \makecell{Clas-\\ses}}
&\multicolumn{1}{c}{ \makecell{Train \\ size}}
&\multicolumn{1}{c}{ \makecell{Train \\ steps}}
&\multicolumn{1}{c}{ \makecell{Batch \\ size}}
&\multicolumn{1}{c}{ \makecell{Learning \\ rate}}
&\multicolumn{1}{c}{ \makecell{Warmup \\ steps}}
&\multicolumn{1}{c}{ \makecell{$\mathbf{W}_{out}$ \\ init}} \\
\midrule
Kinetics 400 & Video & 400 & 215K & \makecell{101K \\ (30 epochs)} & 64 & 0.1 & 2.5 epochs & Zeros \\
Kinetics 600 & Video & 600 & 363K & \makecell{170K \\ (30 epochs)} & 64 & 0.1 & 2.5 epochs & Zeros \\
\makecell{Moments in \\ Time} & Video & 339 & 791K & \makecell{123.6K \\ (10 epochs)} & 64 & 0.25 & 2.5 epochs & Zeros \\
\bottomrule
\end{tabular}
}
\end{table*}

\begin{table*}[h]
\centering
\caption{Set-up for the co-training on audio. Train steps and warmup steps are summed to get the number of train and warmup steps during co-training as we use the ``Weighted'' task sampling method.
}
\label{tab:audiols}
\scalebox{0.9}{
\begin{tabular}{cccccccccc}
\toprule
\multicolumn{1}{c}{ Dataset}
&\multicolumn{1}{c}{ \makecell{Moda-\\lity}}
&\multicolumn{1}{c}{ \makecell{Clas-\\ses}}
&\multicolumn{1}{c}{ \makecell{Train \\ size}}
&\multicolumn{1}{c}{ \makecell{Train \\ steps}}
&\multicolumn{1}{c}{ \makecell{Mixup}}
&\multicolumn{1}{c}{ \makecell{Batch \\ size}}
&\multicolumn{1}{c}{ \makecell{Learning \\ rate}}
&\multicolumn{1}{c}{ \makecell{Warmup \\ steps}}
&\multicolumn{1}{c}{ \makecell{$\mathbf{W}_{out}$ \\ init}} \\
\midrule AS-500k & Audio & 527 & 509K & \makecell{239K \\ (30 epochs)} & 0.5 & 64 & 0.5 & 2.5 epochs & Zeros \\
VGGSound & Audio & 309 & 172K & \makecell{135K \\ (50 epochs)} & 0.3 & 64 & 0.5 & 2.5 epochs & Zeros \\
\bottomrule
\end{tabular}
}
\end{table*}

\end{document}